\documentclass{article}
\pdfoutput=1
\usepackage[preprint]{spconf}
\usepackage{amsmath,graphicx}
\usepackage{subfigure}
\usepackage{amsfonts}
\usepackage[ruled]{algorithm2e}
\usepackage{algorithmic,algorithm2e}
\usepackage{booktabs}
\usepackage{multirow}
\usepackage{color}
\usepackage{floatrow}
\usepackage{comment}
\usepackage{marvosym}
\usepackage{caption}
\usepackage{extarrows}
\usepackage{xcolor}

\def\x{{\mathbf x}}

\title{Hypergraph Transformer for Semi-Supervised Classification}
\name{Zexi Liu\textsuperscript{1}, Bohan Tang\textsuperscript{2}, Ziyuan Ye\textsuperscript{3}, Xiaowen Dong\textsuperscript{2}, Siheng Chen\textsuperscript{1,4}, Yanfeng Wang\textsuperscript{4,1}}
\address{\small{\textsuperscript{1}Shanghai Jiao Tong University, \textsuperscript{2}University of Oxford, \textsuperscript{3}The Hong Kong Polytechnic University, \textsuperscript{4}Shanghai AI Laboratory}}
\ninept 

\newcommand{\R}{\ensuremath{\mathbb{R}}}

\def\Z{\mathbf{Z}}
\def\X{\mathbf{X}}
\def\Y{\mathbf{Y}}
\def\x{\mathbf{x}}
\def\z{\mathbf{z}}
\def\W{\mathbf{W}}
\def\bfQ{\mathbf{Q}}
\def\bfK{\mathbf{K}}
\def\bfV{\mathbf{V}}

\def\y{\mathbf{y}}
\def\hH{\mathbf{H}}

\def\V{\mathcal{V}}
\def\E{\mathcal{E}}
\def\hhH{\mathcal{H}}
\def\G{\mathcal{G}}
\def\L{\mathcal{L}}

\def\O{\mathcal{O}}
\def\A{\mathbf{A}}
\def\D{\mathbf{D}}
\def\P{\mathbf{P}}

\DeclareMathOperator{\h}{\textbf{h}}

\begin{document}
\maketitle
%
\begin{abstract}
Hypergraphs play a pivotal role in the modelling of data featuring higher-order relations involving more than two entities. Hypergraph neural networks emerge as a powerful tool for processing hypergraph-structured data, delivering remarkable performance across various tasks, e.g., hyrgraph node classification. However, these models struggle to capture global structural information due to their reliance on local message passing. To address this challenge, we propose a novel hypergraph learning framework, HyperGraph Transformer (HyperGT). HyperGT uses a Transformer-based neural network architecture to effectively consider global correlations among all nodes and hyperedges. To incorporate local structural information, HyperGT has two distinct designs: i) a positional encoding based on the hypergraph incidence matrix, offering valuable insights into node-node and hyperedge-hyperedge interactions; and ii) a hypergraph structure regularization in the loss function, capturing connectivities between nodes and hyperedges. Through these designs, HyperGT achieves comprehensive hypergraph representation learning by effectively incorporating global interactions while preserving local connectivity patterns. Extensive experiments conducted on real-world hypergraph node classification tasks showcase that HyperGT consistently outperforms existing methods, establishing new state-of-the-art benchmarks. Ablation studies affirm the effectiveness of the individual designs of our model.

\end{abstract}

\section{Introduction}
\label{sec:intro}
Hypergraphs serve as a crucial tool for modelling data with higher-order relationships involving more than two nodes~\cite{bick2023higher,xu2022GroupNet,tang2023learning,xu2023dynamic,tang2023hypergraph}. Hypergraph neural networks leverage message passing over hypergraph structures to enhance node feature learning, leading to remarkable performances in various tasks like node classification~\cite{antelmisurvey}. However, these message-passing-based models suffer from the oversmoothing issue, particularly when increasing the model depth to capture global information~\cite{chen2022preventing,liu2023multi}. Recently, some Transformer-based methods have emerged for hypergraph-structured data~\cite{chien2022you,heo2022hypergraph,li2023hypergraph}. Nonetheless, these methods predominantly use the neighborhood attention to update node features within individual hyperedges,  thereby constraining their ability to fully exploit global information. In summary, the existing literature lacks efficient methods for exploiting the global information inherent in hypergraph-structured data. 

To fill this gap, we propose HyperGraph Transformer (HyperGT), a novel learning framework that efficiently incorporates both global and local structural information in the hypergraph. HyperGT consists of three key components. First, the hypergraph attention operation explores correlations among all nodes and hyperedges, enabling efficient capture of global information.
Second, the positional encoding based on the hypergraph incidence matrix integrates local node-node connectivities into the forward propagation of HyperGT. Third, the hypergraph structure regularization in the loss function, which is derived from the hypergraph star-expansion, encourages HyperGT to capture local node-hyperedge interactions.

We extensively evaluate our model on four real-world hypergraph semi-supervised node classification benchmarks. Results are promising: 1) HyperGT achieves state-of-the-art (SOTA) performance across all datasets, notably surpassing previous SOTA hypergraph neural networks by nearly $3\%$ in classification accuracy on the Walmart dataset; and 2) Ablation studies unequivocally demonstrate the effectiveness of each of three proposed components, as an example: the proposed positional encoding mechanism and structure regularization boost the classification accuracy of a vanilla Transformer by approximately $25\%$ on the Walmart dataset.

\begin{figure}[t] 
\centering
\includegraphics[width=.85\textwidth]{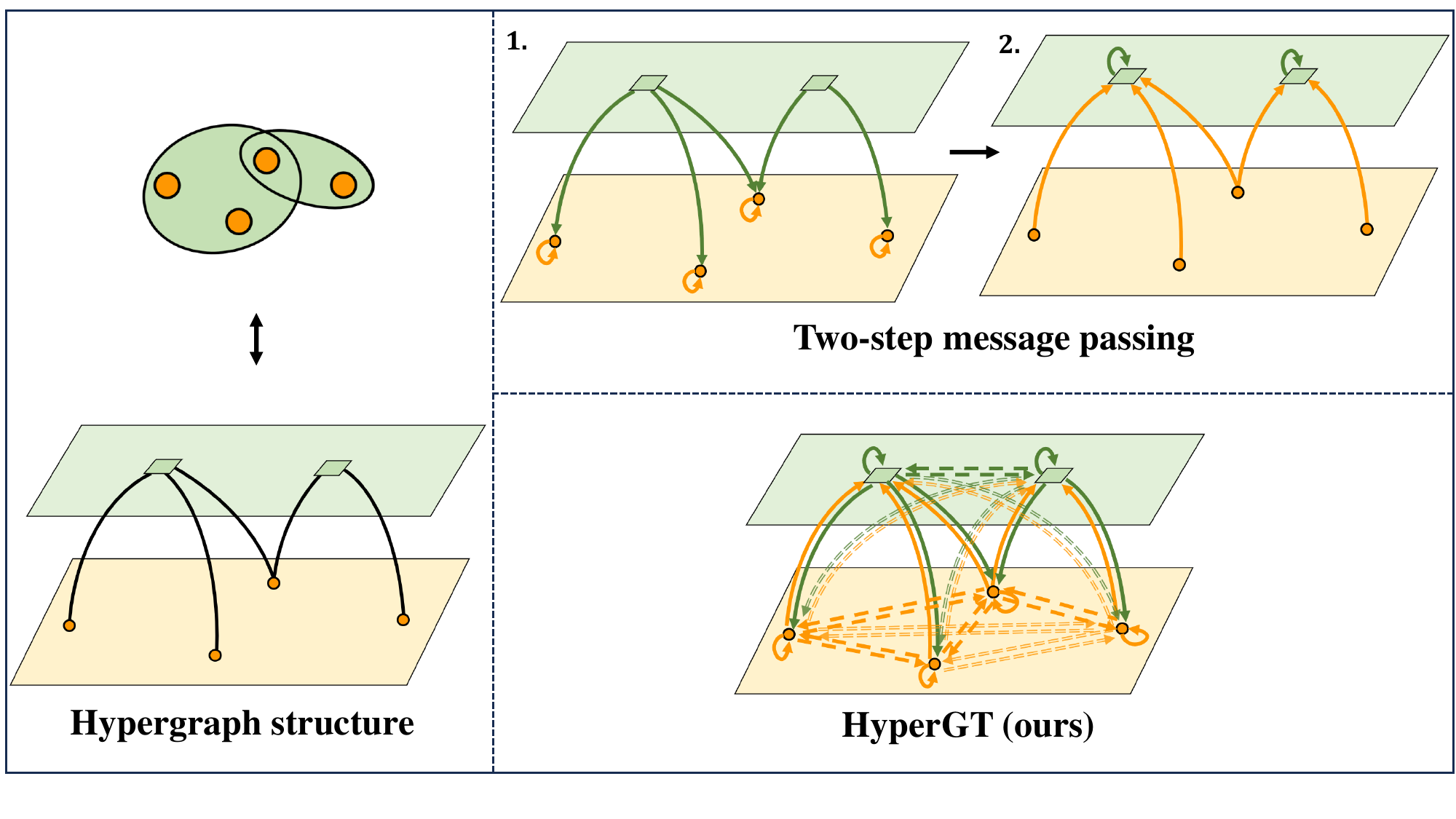}
\vskip -0.05in
\caption{Left: A hypergraph and the associated hypergraph star-expansion. Hypergraph star-expansion fully preserves the structural information of the hypergraph. Right: Illustration of the previous message-passing method and our HyperGT framework for the hypergraph depicted on the left. We build interactions among all nodes and hyperedges by hypergraph attention.}
\label{fig:hypergt}
\vskip -0.1in
\end{figure}

\begin{figure*}[t] 
\centering
\includegraphics[width=.9\textwidth]{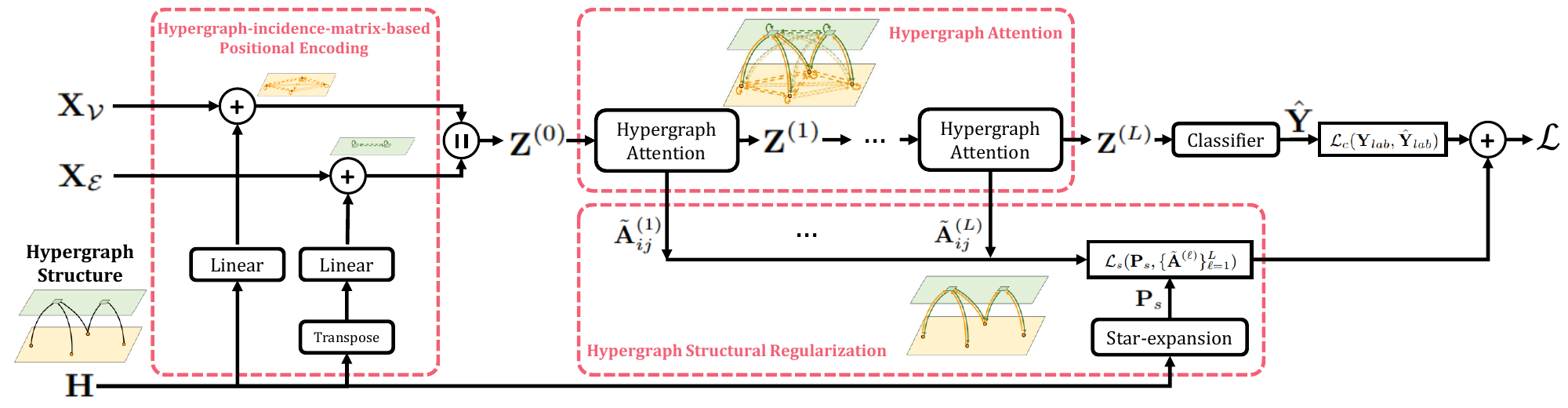}
\caption{System architecture of HyperGT, taking node features $\X_\V$, hyperedge features $\X_\E$, and the hypergraph incidence matrix $\hH$ as inputs. There are three key components: 1) Positional encoding using the hypergraph incidence matrix for node-node and hyperedge-hyperedge interactions; 2) Hypergraph attention module creating connections from one node/hyperedge to any other nodes/hyperedges; and 3) Hypergraph structure regularization for node-hyperedge connectivities.}
\label{fig:dataflow}
\vskip -0.1in
\end{figure*}

\section{Related Work}

\textbf{Hypergraph neural networks.} Numerous studies focus on developing neural networks to process data on hypergraphs. Most existing hypergraph neural networks (HGNNs)~\cite{wang2023equivariant,AllSet,HGNN,HNHN,HCHA,UNIGCN,HyperGCN} follow a two-step message passing paradigm: first, node features are sent to corresponding hyperedges to learn hyperedge embeddings, then the learned hyperedge embeddings are sent back to nodes to learn node embeddings. A recent study~\cite{chen2022preventing} highlights a common challenge encountered in message-passing-based hypergraph neural networks: the oversmoothing issue. This issue leads to increasingly homogeneous node embeddings as the depth of models increases, impeding their capacity to effectively capture valuable global information, such as long-range dependencies, through deeper architectures. Consequently, the current landscape lacks hypergraph neural networks capable of efficiently exploiting the inherent global information present in hypergraph-structured data.

\textbf{Transformer for structured data.} Recently, the Transformer architecture has been adapted to effectively process graph-structured data by incorporating both local and global graph structure information, leading to promising outcomes~\cite{yun2019graph,HOT,GPS,ma2023GraphInductiveBiases}. However, in the field of hypergraph machine learning, the Transformer is merely used to learn local attention mechanisms inside a hyperedge~\cite{chien2022you,heo2022hypergraph,li2023hypergraph}. To our knowledge, our work represents the first effort in harnessing the Transformer architecture to simultaneously capture both local and global information within hypergraph-structured data.

\section{Preliminary}

\textbf{Hypergraphs.} We denote a hypergraph as a triplet $\hhH = \{\V, \E, \hH\}$, where $\V = \{v_1, v_2, \cdots, v_n\}$ is the node set with $|\V|=n$, $\E=\{e_1, e_2, \cdots, e_m\}$ is the hyperedge set with $|\E| = m$, and $\hH \in \{0, 1\}^{n\times m}$ is an incidence matrix in which $\hH_{ij} = 1$ indicates that hyperedge $j$ contains node $i$ and $\hH_{ij} = 0$ otherwise. 

\textbf{Semi-supervised classification on hypergraphs.} We focus on the semi-supervised hypergraph node classification task. Let $\X_\V = [\x_{v_1}^\top, \x_{v_2}^\top, \cdots, \x_{v_n}^\top]^{^\top} \in \R^{n\times d}$ be the node features, which is a matrix that contains $d$-dimensional features, $\V_{lab}$ be the set of labeled nodes with ground truth labels $\Y_{lab}=\{\y_v\}_{v\in\V_{lab}}$, where $\y_{v_i}\in\{0,1\}^{c}$ be a one-hot label, and $\V_{un}=\V\setminus\V_{lab}$ be a set of unlabeled nodes. 
Given node features $\X_\V$, labels $\Y_{lab}$, and the hypergraph structure $\hH$, we aim to classify nodes in $\V_{un}$.

\textbf{Transformer.} In a standard Transformer, the forward propagation begins with a positional encoding mechanism, followed by a sequence of identical Transformer layers, each of which consists of a multi-head self-attention modules (MHSA) and a position-wise fully connected feed-forward network (FFN)~\cite{Attention}. 

The positional encoding mechanism adds the given structural information to the input embeddings. Let $\X\in\R^{n\times d}$ be the input features and $\P\in\R^{n\times d}$ be features embedding the input structural information. The initial structural-augmented embedding is
\begin{equation*}
\setlength{\abovedisplayskip}{.1pt}
\setlength{\belowdisplayskip}{.1pt}
\Z = \X+\P \in\R^{n\times d},
\end{equation*}
whose $i$-th row $\z_{i}\in\R^{d}$ is the initial embedding of the $i$-th input.

The MHSA, a pivotal component for information exchange among entities, update embeddings for each entity.  Let $\Z$ be the input embedding.
Excluding bias terms and multi-head details for simplicity, the self-attention module works as:
\begin{eqnarray*}
\setlength{\abovedisplayskip}{-0.3pt}
\setlength{\belowdisplayskip}{-0.3pt}
\label{trans1}
\mathbf{Q} & = & \Z \mathbf{W}_\mathbf{Q},\quad \mathbf{K}=\Z \mathbf{W}_\mathbf{K},\quad \mathbf{V}=\Z \mathbf{W}_\mathbf{V},
\\
\mathbf{\tilde A} & = & \operatorname{softmax}\left({\frac{\mathbf{Q} \mathbf{K}^{\top}}{\sqrt{d_{k}}}}\right),
\\
\Z' & = &  \mathbf{\tilde A}\mathbf{V},
\label{trans2}
\end{eqnarray*}
where $\mathbf{W}_\mathbf{Q},\mathbf{W}_\mathbf{K}, \mathbf{W}_\mathbf{V}\in\R^{d\times d_k}$ are learnable matrices, and the weight matrix $\mathbf{\tilde A}\in\R^{n\times n}$ is formed from the similarity calculation between any pair of inputs to update their features. 

The FFN after the MHSA serves as a pointwise non-linear transformation for $\Z'$. Here we omit the details for brevity.

\section{Hypergraph Transformer}
In this section, we introduce our HyperGraph Transformer (HyperGT) tailored for the node classification task, as illustrated in Fig.~\ref{fig:dataflow}. Leveraging both node and hyperedge features, we enhance these features by incorporating node-node and hyperedge-hyperedge interactions through a positional encoding mechanism driven by the hypergraph incidence matrix, thus generating input embeddings (Section~\ref{sec:pe}). These embeddings are thereon fed in the hypergraph attention module, which establishes dense connections between nodes and hyperedges (Section~\ref{sec:att}). Finally, we compute the training loss $\L$ by combining the traditional supervised classification loss $\L_c$ with a hypergraph structure regularization loss $\L_s$, adept at capturing node-hyperedge interrelations (Section~\ref{sec:reg} and Section~\ref{sec:train}).

\subsection{Hypergraph incidence matrix based positional encoding}
\label{sec:pe}
HyperGT simultaneously incorporates node features and hyperedge features as the input. Let $\X_\E = [\x_{e_1}^\top, \x_{e_2}^\top, \cdots, \x_{e_m}^\top]^{^\top} \in \R^{m\times d}$ denote the hyperedge features where we let the dimension of hyperedge features be consistent with that of node features. 
If hyperedge features are not available, to make node and hyperedge features correlated, we initialize hyperedge features by computing the average of features of nodes within that hyperedge. 
The input features are:
\begin{equation}
\setlength{\abovedisplayskip}{.1pt}
\setlength{\belowdisplayskip}{.1pt}
\label{eq:input}
\X=\begin{bmatrix}
  \X_\V \\
  \X_\E
\end{bmatrix},
\end{equation}
where $\X\in\R^{(n+m)\times d}$ means each node and each hyperedge serves as the input of the global attention module.

The functionality of the positional encoding is to provide structural insights for nodes and hyperedges, enabling our model to become aware of the hypergraph structure. To achieve this, we inject structural information to the input features defined in Eq.~(\ref{eq:input}). Specifically, since the hypergraph incidence matrix $\hH$ inherently embodies the hypergraph structure itself, we leverage it to design learnable positional encoding for each node and hyperedge. 

Let $\P_\V$ and $\P_\E$ denote the positional encodings for nodes and hyperedges, respectively. Then, our positional encoding works as

\begin{equation*}
\setlength{\abovedisplayskip}{.1pt}
\setlength{\belowdisplayskip}{.1pt}
\P_\V=\hH \mathbf{W}_\V, \P_\E=\hH^\top \mathbf{W}_\E,
\end{equation*}
where $\mathbf{W}_\V\in\R^{n\times d}, \mathbf{W}_\E\in\R^{m\times d}$ embeds the nodes and hyperedges structure via a learnable linear projection to generate positional encodings. Then, it is added to the input features directly:
\begin{equation*}
\Z^{(0)}= \X + \P =
\begin{bmatrix}
  \X_\V+\P_\V \\
  \X_\E+\P_\E
\end{bmatrix}=
\begin{bmatrix}
\begin{aligned}
&\X_\V+\hH \mathbf{W}_\V \\
&\X_\E+\hH^\top \mathbf{W}_\E
\end{aligned}
\end{bmatrix}
\in\R^{(n+m)\times d}.
\end{equation*}
By using the positional encoding, our model can effectively leverage the hypergraph structural information.

To further clarify the advantage of our proposed positional encoding, we show that it can represent the local structure of the hypergraph. Define the structure information of each node to be the hyperedges that contain this node, and the structure information of each hyperedge as the nodes contained within this hyperedge. Then, we have the following result:

\noindent \textbf{Theorem 1.} Let $\mathbf{p}_u$, $\mathbf{p}_v$ be the positional encoding of nodes $u$ and $v$, respectively. Then, $\lVert\mathbf{p}_u-\mathbf{p}_v\rVert_2\leq C \sqrt{N_{e}}$,
where $C$ is a constant and $N_{e}$ is the number of hyperedges only with either node $u$ or $v$.

\noindent Proof. 
\begin{eqnarray}
\label{inequality}
\setlength{\abovedisplayskip}{.1pt}
\setlength{\belowdisplayskip}{.1pt}
\lVert \mathbf{p}_u-\mathbf{p}_v\rVert_2&=& \lVert \h_uW_\V-\h_vW_\V \rVert_2\nonumber\\  &=&\lVert(\h_u-\h_v)W_\V\rVert_2\nonumber\\ 
&\leq& \lVert W_\V\rVert_2\cdot\lVert(\h_u-\h_v)\rVert_2 \nonumber\\ 
&=&\sigma_{max}(W_\V)\sqrt{|\{e|u\notin e, v\in e \cap u\in e, v\notin e\}|} \nonumber\\ 
&=&C\sqrt{N_{e}},
\end{eqnarray}
where $\h_u,\h_v$ are the $u$-th and $v$-th row vectors of $\hH$, $\lVert \cdot \rVert_2$ denotes the 2-norm of a vector or a matrix, and $\sigma_{max}(\cdot)$ denotes the largest singular value of a matrix. 
From Eq. (\ref{inequality}), we know that
the distance between the positional encodings of the two nodes is bounded by the number of hyperedges that only contain either node $u$ or $v$, while the latter represents the similarity between the structural information of node $u$ and $v$. We have similar results for hyperedges as well. 

In summary, positional encoding enhances structural information utilization in our model for nodes and hyperedges.

\subsection{Hypergraph attention}
\label{sec:att}
The functionality of the hypergraph attention is to explore correlations among all nodes and hyperedges, capturing  global information. Mathematically, let  $\Z^{(\ell)}$ be the feature embedding at the $\ell$-th hypergraph attention layer with $\z_{i}^{\ell}$ the hidden representation of instance $i$ and $\tilde{\mathbf{A}}^{(\ell)} \in \R^{(n+m)\times (n+m)}$ be the softmax attention matrix at the $\ell$-th layer. The ($i$, $j$)-element of $\tilde{\mathbf{A}}^{(\ell)}$ is obtained as
\begin{equation*}
\label{eqn:element-wise_att}
\tilde{\mathbf{A}}_{ij}^{(\ell)}=\frac{\exp((\z_i^{(\ell)}\W_\bfQ^{(\ell)})({\z}_j^{(\ell)}\W_\bfK^{(\ell)})^\top)}{\sum_{k=1}^{n+m}\exp((\z_i^{(\ell)}\W_\bfQ^{(\ell)})({\z}_k^{(\ell)}\W_\bfK^{(\ell)})^\top)},
\end{equation*}
where $\W_\bfQ^{(\ell)}, \W_\bfK^{(\ell)}, \W_\bfV^{(l)}$ are three learnable matrices at the $l$-th layer. After attention is calculated between any two inputs, the representation of the $i$-th instance is updated as
\begin{equation*}
\setlength{\abovedisplayskip}{.1pt}
\setlength{\belowdisplayskip}{.1pt}
\z_i^{(\ell+1)}=\sum_{j=1}^{n+m}\tilde {\mathbf{A}} _{ij}^{(\ell)}\cdot(\z_j^{(\ell)}\W_\bfV^{(\ell)}).
\end{equation*}
This creates a dense connection from one node/hyperedge to any other nodes/hyperedges in only a single step. 

This design considers a comprehensive attention among nodes and hyperedges. It has two distinct differences from previous works. First, in previous works, the interaction between two nodes is established along the node-hyperedge-node pathways. Thus, information is propagated slowly through the same local kernel, causing the oversmoothing issue. In comparison, we enable the pairwise interactions between all instances (nodes and hyperedges), which makes information propagation more efficient. Second, previous methods do not consider interactions between hyperedges, while we establish the possible attentions between hyperedges to get better hyperedge embeddings, further crafting better node representations.

\subsection{Hypergraph structure regularization}
\label{sec:reg}

To further encourage the model to leverage local node-hyperedge interaction, we introduce a hypergraph structure regularization. The key idea is to utilize the node-hyperedge connection prior to guide the training of the attention matrix.

To obtain a supervision for the softmax attention matrix, we build a probabilistic transition matrix  based on the star-expansion structure of the hypergraph. Mathematically, let $\G=\{\V_s,\E_s,\A_s\}$ be the hypergraph star-expansion, where the vertex set is augmented with hyperedges $\V_s=\V\cup\E$, and vertices and hyperedges are connected by their incident relations $\E_s=\{(v,e)\mid v\in e, v\in\V,e\in\E\}$. The adjacency matrix $\A_s\in\{0,1\}^{(n+m)\times(n+m)}$ embeds this star-expansion graph structure:
\begin{equation*}
\A_s=\left[\begin{array}{c c} 0_{|\V|}&\hH\\ 
                            \hH^{\top}&0_{|\E|}\end{array}\right].
\end{equation*}
Notably, $\A_s$ is uniquely associated with $\hH$, and one can reconstruct the original hypergraph from the star-expansion structure without any information loss. Therefore, the star-expansion structure of the hypergraph fully preserves the structural information of the hypergraph (i.e., $\hH$). The corresponding probabilistic transition matrix is $\P_s = \D_s^{-1} \A_s \in \R^{(n+m)\times(n+m)}$, where $\D_s$ is a diagonal matrix with $(\D_s)_{ii} = \sum_j (\A_s)_{ij}$. The $(i,j)$-th element in $\P_s$ represents the probability that the $i$-th instance transits to the $j$-th instance.

This transition matrix $\P_s$ is an appropriate supervision for $\tilde\A^{(\ell)}$ for three reasons: i) $\P_s$ can reflect the complete relations between nodes and hyperedges; ii) through the star-expansion, $\P_s$ and $\tilde\A^{(\ell)}$ have the same dimension; and iii) both $\P_s$ and $\tilde\A^{(\ell)}$ have probabilistic meanings. Then, the hypergraph structure loss is defined based on the cross-entropy between $\P_s$ and $\tilde\A^{(\ell)}$:
\begin{eqnarray}
\setlength{\abovedisplayskip}{.1pt}
\setlength{\belowdisplayskip}{.1pt}
\label{Le}
\L_{s}(\P_s,\{\tilde{\mathbf{A}}^{(\ell)}\}_{\ell=1}^L)
&\! =\! &  -\frac{1}{(n+m) L}\sum_{\ell=1}^L \sum_{i,j=1}^{n+m} (\P_s)_{i,j} \log \tilde{\mathbf{A}}_{i j}^{(\ell)}
\nonumber \\ \nonumber
&\! =\! & -\frac{1}{(n+m) L}\sum_{l=1}^{L}\sum_{(i,j)\in\E_s}\frac{1}{d_{i}}\log \tilde{\mathbf{A}}_{i j}^{(\ell)},
\end{eqnarray}
where $L$ is the number of layers in HyperGT and $d_i$ is the degree of instance $i$ in the star-expansion structure of the hypergraph.

\subsection{Training strategy} 
\label{sec:train}
For node classification, given training labels $\Y_{lab}=\{\y_u\}_{u\in\V_{lab}}$, where $\V_{lab}$ denotes the set of labeled nodes, we train our model with the cross-entropy loss function: 
\begin{equation*}
\setlength{\abovedisplayskip}{.1pt}
\setlength{\belowdisplayskip}{.1pt}
\L_{c}({\bf Y}_{lab},\hat{{\bf Y}}_{lab})= -\frac{1}{|\V_{lab}|}\sum_{u\in\V_{lab}}\y_u\log(\hat{\y}_u^\top),
\end{equation*}
where $\log(\cdot)$ is an element-wise logarithmic function. Considering our hypergraph structure loss, the final objective is: 
$\L = \L_c +\lambda\L_s$, where $\lambda$ is a coefficient to balance the two losses. This learning objective is minimized to update trainable parameters in HyperGT.

\begin{table}[t]
\begin{center}\tiny
\caption{Properties of datasets.}
\vskip -0.1in
\label{property}\resizebox{.9\columnwidth}{!}{
\begin{tabular}{ccccc}
\hline
 &Congress & Senate & Walmart & House 
\\
\hline
$|\V|$ & 1718&  282&  88860& 1290
\\
$|\E|$ & 83105&  315&  69906& 341
\\
\# features & 100&  100&  100& 100
\\
\# class & 2&  2&  11& 2
\\
\hline
\end{tabular}}
\end{center}
\vskip -0.1in
\end{table}

\begin{table}[t]
\begin{center}\normalsize
\caption{Results for the tested datasets: Mean accuracy (\%) ± standard deviation. Experiments show that HyperGT outperforms existing state-of-the-art hypergraph neural networks.}
\vskip -0.1in
\label{results}\resizebox{.9\columnwidth}{!}{
\begin{tabular}{ccccc}
\hline
 &Congress & Senate & Walmart & House \\
\hline
HGNN & 91.26 ± 1.15&48.59 ± 4.52&62.00 ± 0.24&61.39 ± 2.96\\
HCHA & 90.43 ± 1.20&48.62 ± 4.41&62.35 ± 0.26&61.36 ± 2.53\\
HNHN &53.35 ± 1.45& 50.93 ± 6.33& 47.18 ± 0.35& 67.80 ± 2.59\\
HyperGCN&55.12 ± 1.96&42.45 ± 3.67& 44.74 ± 2.81& 48.32 ± 2.93\\
UniGCNII&94.81 ± 0.81& 49.30 ± 4.25& 54.45 ± 0.37& 67.25 ± 2.57\\
HyperND& 74.63 ± 3.62& 52.82 ± 3.20& 38.10 ± 3.86& 51.70 ± 3.37\\
AllDeepSets&91.80 ± 1.53& 48.17 ± 5.67& 64.55 ± 0.33& 67.82 ± 2.40\\
AllSetTransformer& 92.16 ± 1.05& 51.83 ± 5.22& 65.46 ± 0.25& 69.33 ± 2.20\\
ED-HNN&95.00 ± 0.99&64.79 ± 5.14&66.91 ± 0.41&72.45 ± 2.28\\

HyperGT(Ours)&\textbf{95.23 ± 0.73}&\textbf{65.49 ± 5.11}&\textbf{69.83 ± 0.39}&\textbf{74.55 ± 1.99}\\
\hline
\end{tabular}}
\end{center}
\vskip -0.2in
\end{table}

\section{Experiments}
\subsection{Experimental settings}

\textbf{Datasets.} We use four datasets from previous hypergraph neural networks research\cite{wang2023equivariant}. In the House dataset, nodes represent US House of Representatives members, with hyperedges corresponding to committee memberships. Node labels indicate the political party. In Walmart, hyperedges represent co-purchased product sets, and nodes are products labeled with categories. In Congress and Senate dataset, nodes are US Congresspersons labeled with political party affiliation, and hyperedges are formed by bill sponsors and co-sponsors from both chambers. Since these datasets lack node attributes, we follow Chien et al. \cite{AllSet} to generate node features from label-dependent Gaussian distribution, where the standard deviation of the added Gaussian features is set as 1.0. The partial statistics of all datasets used are provided in Table \ref{property}.

\textbf{Baselines.}
We compare our method with top-performing models on these benchmarks, including HGNN~\cite{HGNN}, HCHA~\cite{HCHA}, HNHN~\cite{HNHN}, HyperGCN~\cite{HyperGCN}, UniGCNII~\cite{UNIGCN}, AllDeepSets~\cite{AllSet}, AllSetTransformer ~\cite{AllSet}, HyperND~\cite{tudisco2021nonlinear}, and ED-HNN~\cite{wang2023equivariant}. All the hyperparameters for baselines are followed from \cite{wang2023equivariant}. 

\textbf{Implementation details.}
 Our model is built upon the implementation of Nodeformer \cite{Nodeformer}, which leverages a kernerlized gumbel-softmax operator to reduce the attention computation comlexity to linearity. We identified $\lambda$’s optimal value via grid-search. Regarding the hyperparameters of the $\O(N)$ Transformer, we set them to be consistent with NodeFormer. We randomly split the data (node indices) into training/validation/test samples using 50\%/25\%/25\% splitting percentages as in Chien et al. \cite{AllSet}. We report the average prediction accuracy over ten random splits as the evaluation metric. All experiments are conducted on an RTX 3090 utilizing the PyTorch framework. Our code is available at: https://github.com/zeroxleo/HyperGT.

\subsection{Results and analysis}
\textbf{Comparison with SOTA hypergraph neural networks.} The results of our experiments on real-world hypergraph node classification benchmarks are presented in Table \ref{results}. In comparison to previous hypergraph neural networks, our proposed HyperGT demonstrates superior classification accuracy across all datasets. A key distinguishing feature of HyperGT is its incorporation of both global and local hypergraph structural information, setting it apart from prior methods that solely focus on local structural information. These results show the importance of leveraging global information for designing more effective models in processing hypergraph-structured data. We also conducted experiments on the Walmart dataset to obtain the inference speed of popular HGNNs and HyperGT. Results are presented in Table \ref{time}, indicating that HyperGT maintains competitive inference speeds, aligning with its theoretically low linear complexity, while also preserving global attention interactions.

\textbf{Ablation studies.} We perform a series of ablation studies on the importance of designs in our proposed HyperGT on the walmart dataset. The ablation results are presented in Table \ref{ablation_study}. 
We see that the use of positional encodings for nodes leads to a significant improvement, which shows that the model effectively captures the hypergraph structural information of nodes. Positional encodings of hyperedges also contribute to positive effects. With the addition of regularization, the performance is further improved, which indicates that this regularizer on hypergraph structure helps the Transformer in learning the specific connections between nodes and hyperedges. All of the components designed are helpful to our Transformer architecture for modelling hypergraph data and help to leverage useful information from input hypergraphs.

\begin{table}[t]
\begin{center}
\caption{Ablation study on the Walmart dataset. Results demonstrate the effectiveness of each proposed components.}
\vskip -0.1in
\label{ablation_study}\resizebox{.9\columnwidth}{!}{
\begin{tabular}{ccccc}

\hline

node PE&hyperedge PE&structure regularization& ACC\\
\hline  
    -     &    -     &    -     &  45.67   \\
\hline
\checkmark&    -     &    -     &  66.51       \\
\hline
\checkmark&\checkmark&    -     &  67.63      \\
\hline
\checkmark&\checkmark&\checkmark&    69.83    \\
\hline
\end{tabular}}
\end{center}
\vskip -0.1in
\end{table}

\begin{table}[t]
\begin{center}
\caption{Inference speeds (ms/run, mean of 1000 runs) on the Walmart dataset. Results show the competitive inference speed of HyperGT.}
\vskip -0.1in
\label{time}\resizebox{.9\columnwidth}{!}{
\begin{tabular}{cccccc}

\hline

HGNN& HCHA& UniGCNII& AllDeepset& AllSetTransformer& HyperGT\\
\hline  
20.301& 20.942& 25.015& 32.603& 62.788 & 24.882\\

\hline
\end{tabular}}
\end{center}
\vskip -0.3in
\end{table}

\section{Conclusion}

In this paper, we present HyperGraph Transformer (HyperGT), a novel learning framework tailored for hypergraph-structured data. HyperGT effectively tackles the challenge of simultaneously capturing global and local structural information within hypergraphs by integrating a Transformer-based architecture with hypergraph-specific components including a hypergraph-incidence-matrix-based positional encoding and a hypergraph structure regularization. Our experiments show that HyperGT outperforms existing state-of-the-art hypergraph neural networks, making it a valuable tool for various applications involving hypergraph-structured data. We leave the development of a theoretical framework to uncover the mechanisms behind the improved performance resulting from the inclusion of global structural information for future research.

\section*{Acknowledgment}
This research is supported by NSFC under Grant 62171276 and 62211530109, as well as the Science and Technology Commission of Shanghai Municipal under Grant 21511100900, 22511106101, and 22DZ2229005. X.D. acknowledges support from the Oxford-Man Institute of Quantitative Finance, the EPSRC (EP/T023333/1), and the Royal Society (IEC\textbackslash NSFC\textbackslash 211188).

\bibliographystyle{IEEEbib}
\bibliography{refs}

\end{document}